# How Psychological Learning Paradigms Shaped and Constrained Artificial Intelligence

Alex Anvi Eponon[1*], Ildar Batyrshin[1], Christian E. Maldonado-Sifuentes[1], Grigori Sidorov[1]

[1*]Centro de Investigación en Computación (CIC), Instituto Politécnico Nacional (IPN), Mexico City, Mexico.

*Corresponding author(s). E-mail(s): aeponon2023@cic.ipn.mx;
Contributing authors: batyr1@cic.ipn.mx;
cmaldonandos2018@cic.ipn.mx; sidorov@cic.ipn.mx;

**Abstract**

Current artificial intelligence systems struggle with systematic compositional reasoning: the capacity to recombine known components in novel configurations. This paper argues that the failure is architectural, not merely a matter of scale or training data, and that its origins lie in the psychological learning theories from which AI paradigms were derived. The argument proceeds in three stages. First, drawing on the systematicity debate in cognitive science and on the demonstration of Aizawa that neither connectionism nor classicism can make systematicity a structural consequence of the architecture, the paper establishes that the corrective techniques proliferating in modern AI, from chain-of-thought prompting to alignment through human feedback, function as auxiliary hypotheses. These hypotheses address symptoms without resolving the underlying architectural indifference to systematicity. Second, the paper traces the genealogy from psychological learning theory to AI methodology. It shows that behaviourism, cognitivism, and constructivism each bequeathed a specific structural limitation to the AI paradigm it inspired: the exclusion of internal structure, the opacity of representation, and the absence of formal construction operators. A cross-cultural reappraisal of rote learning reveals a further underexploited pathway. Third, the paper introduces ReSynth, a trimodular conceptual framework that proposes the principled separation of reasoning, identity, and memory. The framework offers a path toward architectures in which systematic behaviour is a structural consequence of design rather than a correction applied after the fact.

# 1 Introduction

Large language models can summarise legal documents, generate working code, and pass medical licensing examinations. Yet when presented with a task that requires recombining familiar components in a configuration not seen during training, the same systems can fail with striking regularity. Math-specialised models that solve competition-level problems cannot chain two elementary grade-school operations [35]. Models that parse complex sentences lose coherence when the same words appear in unfamiliar arrangements [36]. Systems that achieve near-perfect accuracy on standard benchmarks collapse when tested on formally verified novel combinations [51]. The pattern is consistent: current AI architectures can reproduce and interpolate, but they struggle to systematically recombine.

Queloz [6] has argued that this limitation has particular consequences for normative domains: where the truth is asystematic, where true statements do not form a consistent, coherent whole, large language models cannot leverage systematicity to fill gaps in their training data, and we therefore cannot rely on them for practical deliberation. The subsequent exchange with Leuenberger [7] and the reply of Queloz [8] further explored how the "spurious systematicity" imposed by corrective techniques conceals rather than resolves the underlying problem. This paper argues that the failure Queloz identified in normative domains is a symptom of a deeper architectural condition. The dominant paradigms of artificial intelligence struggle with systematicity not because particular domains resist systematic modelling, but because the architectures themselves are structurally indifferent to systematicity, and this indifference is inherited from the current understanding, or more precisely the numerical representation of original psychological learning theories on which those architectures were built.

The claim advanced here is therefore that the compositional reasoning failures documented across current AI systems are neither a temporary limitation to be resolved by scale nor an engineering problem amenable to incremental refinement. They are an architectural constraint, and its origins are deeper than is commonly recognised. The dominant paradigms of artificial intelligence were not designed from first principles; they were borrowed from psychology. Reinforcement learning implements behaviourist assumptions about stimulus-response conditioning. Deep learning implements cognitivist assumptions about internal representation. Integrative approaches implement constructivist assumptions about knowledge assembly. Each paradigm inherited not only the explanatory power of its psychological ancestor but also its structural limitations. Partial solutions exist: meta-learning algorithms [18, 76], continual learning methods [32], and modular architectures [33] represent genuine advances. But as this paper will argue, they function as corrective overlays applied to architectures whose base theories do not structurally require the adaptive behaviour they produce.

The paper develops this argument in three stages. Section 2 establishes the problem: why current AI architectures are structurally indifferent to systematicity, drawing on the systematicity debate in cognitive science and particularly on the analysis of Aizawa [39], who demonstrated that neither connectionism nor classicism can make systematic behaviour an architectural necessity. This section also shows how the corrective techniques that Queloz and Leuenberger discuss, from chain-of-thought prompting to alignment through reinforcement learning from human feedback, function as what Aizawa calls auxiliary hypotheses: additions that the base theory does not necessitate, analogous to the Ptolemaic astronomer's ad hoc colinearity hypothesis. Section 3 explains the origin of the problem: a genealogy from psychological learning theories to AI methodologies that identifies the specific assumptions each paradigm carried forward and the specific limitations that came with them, including a cross-cultural reappraisal of rote learning that reveals an underexploited pathway. Section 4 proposes an escape: the ReSynth framework, which separates reasoning, purpose, and memory into constitutively independent components, and argues that systematic behaviour, in the sense demanded by the systematicity debate, becomes a structural consequence of this separation rather than an accidental product of auxiliary hypotheses.

The argument thus contributes to an emerging conversation in the philosophy of technology about what current AI systems can and cannot achieve in principle, and about the theoretical commitments, often unexamined, that constrain their design. If the corrective techniques proliferating across modern AI are, as we argue, structurally analogous to Ptolemaic epicycles, then understanding why requires tracing their origins beyond engineering choices to the psychological theories that shaped the architectures in the first place. The practical stakes are considerable: the question of whether systematic reasoning can be engineered into existing paradigms or whether it requires a different architectural foundation has direct implications for how we develop, evaluate, and govern AI systems.

# 2 The Systematicity Problem in Current AI

The difficulty that current AI systems face with compositional reasoning is not an edge case. It is pervasive, empirically well documented, and, as this section demonstrates, a consequence of architectural choices that possibly no amount of engineering refinement within the existing paradigms can resolve.

## 2.1 Empirical Evidence of Architectural Failure

Recent empirical work confirms these limitations with striking specificity. In their research, Kirkpatrick et al. [32] showed that conventional neural networks exhibit catastrophic forgetting while learning tasks in sequence, primarily owing to the fact that the shared parameter space cannot support old representations during new learning. It is not an exception but a structural effect resulting from the entanglement of representations. The architecture based on separated modular systems has been found to work efficiently in tackling the above problem, as evidenced by a study by Ellefsen et al. [33], which indicates that neural networks evolve with a lower propensity for catastrophic forgetting due to modularity. Aly and Dugan [34] showed that dynamic information balancing across modular networks outperforms monolithic architectures with equivalent capacity on continual learning benchmarks.

The compositional reasoning failures of large language models provide further evidence. Hosseini et al. [35] showed that even math-specialised LLMs capable of solving competition-level problems fail on compositional grade-school mathematics tasks that merely chain two elementary problems together. Three main forms of systematic failure in compositional reasoning have been reported by Li et al. [36]: missing intermediate memory, spurious retrieval from high-frequency distractors, and activation drift between reasoning steps. A comprehensive analysis of LLM scaling limits concluded that likelihood-based training rewards local coherence rather than logical entailment, leading to syntactic rather than semantic generalisation [37]. These are not failures of scale; they are consequences of architectures that compress reasoning, memory, and purpose into shared parameters where each function interferes with the others.

## 2.2 The Systematicity Argument

This diagnosis gains further precision from the systematicity debate in cognitive science. The debate, initiated by Fodor and Pylyshyn [38], concerns whether cognitive architecture must support *compositional* representations which are structured wholes built from atomic constituents, and whether having such representations is sufficient to guarantee *systematic* behaviour: the property that the capacity to entertain one thought (e.g., that John loves Mary) entails the capacity to entertain its systematic variants (e.g., that Mary loves John). Fodor and Pylyshyn argued that connectionist architectures, defined minimally as networks of nodes with weighted connections, cannot make systematicity a necessary consequence of the architecture. A network can be wired to produce a systematic set of representations, but it can equally be wired to produce a non-systematic set; the base theory is, as Fodor and McLaughlin put it, "absolutely indifferent" between these options [71].

Aizawa [39] deepened this challenge in two decisive ways. First, he demonstrated that the same problem afflicts *classicism*, the hypothesis that cognition operates over a compositional language of thought (LOT). This is how the argument is presented. Consider a Turing machine that computes over a compositional alphabet {John, Mary, Jane, loves, hates, fears}. The programme of the machine might allow it to write 'John loves Mary', 'John loves Jane', and 'Mary loves Jane', while disallowing 'Mary loves John', 'Jane loves John', and 'Jane loves Mary'. Such a machine displays syntactic and semantic compositionality: it constructs complex representations from atomic parts, and the meanings of those representations are determined by the meanings of the parts and their mode of combination. Yet the capacity of the machine to represent John's loving Mary does not bring with it the capacity to represent Mary's loving John. Compositionality, Aizawa concludes, does not entail systematicity. The claim is not merely that compositionality fails to entail some *particular* systematic relation, it is the stronger claim that compositionality does not entail that there be *any* systematic relations between possible representations at all.

This result translates directly to contemporary deep learning. Transformer architectures are, in a meaningful sense, compositional: attention mechanisms decompose inputs into contextualised token representations, which are then combined through multi-layer processing to produce structured outputs. But this compositional processing does not guarantee systematic recombination. The empirical evidence confirms this: the failures documented by Hosseini et al. [35] and Li et al. [36] are precisely the symptoms one would expect from an architecture that permits but does not require systematic composition.

## 2.3 The $P*$ Problem: Why Corrective Overlays Do Not Suffice

The second decisive contribution of Aizawa was to show that the natural response to this problem, which is adding an auxiliary hypothesis to the base theory, does not resolve it. A connectionist might propose some property $P*$ such that any network satisfying $P*$ produces systematic outputs. But, Aizawa argues, this merely puts the connectionist in the same position as the Ptolemaic astronomer. Ptolemy could account for the fact that Mercury and Venus never appear in opposition to the Sun by adding the hypothesis that their deferents are locked to the Sun. The hypothesis saves the phenomena, but it saves them *in the wrong way* : the basic Ptolemaic framework is equally consistent with the deferents being non-colinear, so the colinearity hypothesis is an arbitrary addition that the base theory does not necessitate. By contrast, the Copernican heliocentric model entails the bounded elongation of Mercury and Venus as a structural consequence, without any additional hypothesis. In the analysis of Aizawa, conjoining $P*$ with bare connectionism is structurally identical to conjoining the colinearity hypothesis with bare geocentrism: the base theory remains indifferent to the phenomenon, and the auxiliary hypothesis does the explanatory work that the theory itself should provide.

What could be the relevance to modern AI in practice, is immediate. The proliferation of corrective techniques in contemporary systems where the most common among them are chain-of-thought prompting [40], retrieval-augmented generation [41], alignment through reinforcement learning from human feedback [42], and Constitutional AI [70] can be understood, in the terms of Aizawa, as instances of $P*$: auxiliary properties conjoined with the base connectionist theory to produce behaviour that the theory alone does not guarantee. Each technique addresses a specific failure mode, but none alters the architectural fact that the base theory is indifferent to systematicity. The system does not produce systematic outputs *because of* its architecture, it produces them *despite* its architecture, under the right conditions, with the right corrections, for the right inputs.

A further dimension of his analysis addresses an objection that might be raised against the present argument. One might respond that even if bare compositionality does not entail systematicity, a sufficiently *rich* compositional system, specifically, one whose language is syntactically and semantically recursive, would entail systematic relations among its possible representations. Aizawa considers this "strong LOT" proposal and demonstrates that it, too, fails. A recursive grammar does guarantee that *some* systematic relations will hold (e.g., if the system can represent $S$, it can represent $\neg S$, $\neg\neg S$, and so on). But it does not guarantee the *specific* systematic relations that characterise cognition: that the capacity to think that John loves Mary entails the capacity to think that Mary loves John. Moreover, Aizawa shows that for any *finite* set of sentences, a recursive grammar can be constructed that generates some members of the set but not others, producing gaps in the systematic relations that the theory is supposed to explain. This result maps onto the scaling debate in AI: adding more parameters (more recursive capacity) to a language model does not resolve compositional failures, because the recursive structure of the model does not entail that its finite outputs will exhibit the specific systematic relations that robust generalisation requires. The comprehensive analysis of LLM scaling limits by Mohsin et al. [37] confirms this: likelihood training rewards local coherence rather than logical entailment regardless of scale.

The SCAN benchmark [51] provides rigorous empirical confirmation of Aizawa's theoretical diagnosis applied to modern systems: when neural networks are tested on novel compositional combinations of familiar primitives, accuracy drops dramatically, indicating that apparent systematic behaviour on established benchmarks reflects memorisation of $P^*$-shaped corrections rather than genuine compositional generalisation.

## 2.4 Convergent Evidence

The framework proposed by Chollet [43] for measuring intelligence provides independent convergent support. Chollet defines intelligence not as task performance but as skill-acquisition efficiency: the ability to handle novel tasks drawn from a broad scope, using limited prior experience. The ARC-AGI benchmark [43] was designed to operationalise this criterion, and the persistent difficulty that large-scale neural models face on ARC tasks, despite their success on conventional benchmarks, is precisely what the present analysis predicts. ARC tasks require the identification and application of abstract relational constraints to novel configurations: the kind of compositional, systematic reasoning that, as Aizawa demonstrated, connectionist architectures can produce only through auxiliary hypotheses rather than as an architectural necessity. Marcus [48] has made a related point from a different direction, arguing that robust AI requires the integration of systematic compositional structure with learned representations, and that purely data-driven approaches face fundamental limitations in achieving this integration.

Thus, current AI architectures are structurally most of the time indifferent to systematicity, and the corrective techniques that have proliferated to address this indifference function as auxiliary hypotheses rather than architectural resolutions. The question that naturally arises is: *why* are the architectures structured this way in the first place? The answer, as the next section argues, lies not in engineering choices but in the psychological theories from which these architectures were derived.

# 3 The Psychological Inheritance

The architectural limitations diagnosed in the previous section are not accidental. They are inherited. The dominant paradigms of artificial intelligence were modelled on theories of human and animal learning, and each paradigm carried forward not only the explanatory power of its psychological ancestor but also its structural constraints. This section traces the genealogy with the precision it requires, identifying the specific assumptions that were transmitted and the specific limitations that came with them (see Fig. 1).

Learning, in the broadest sense, is the process of acquiring new understanding, knowledge, or skills that improve the capacity of the learner to act in current and future situations [1]. This definition, simple as it appears, conceals a profound disagreement about what learning actually involves, a disagreement that has shaped the entire trajectory of artificial intelligence.

## 3.1 Behaviorism and Reinforcement Learning

Behaviourism, the dominant psychological paradigm of the early-to-mid twentieth century, held that learning is best understood as the acquisition of new behaviours through environmental conditioning [2]. The approach was deliberately agnostic about internal mental states: what matters is the relationship between stimulus and response, reinforced through reward and punishment. The mind of the learner is treated as a black box; only observable outputs count as evidence of learning [3].

The strengths of this approach are well known: it provides a clean, empirically grounded framework for studying learned behaviour, and it produced some of the most replicable findings in the history of psychology, from the conditioning experiments of Pavlov to the operant conditioning paradigms of Skinner [4]. Its limitations are equally well documented. Behaviourism cannot account for language acquisition [5], cannot explain how learners generalise beyond their conditioning history, and deliberately excludes the internal representational structure that, as later theories would argue, is precisely what makes human learning flexible and compositional [9].

The mapping to artificial intelligence is direct. Reinforcement learning (RL) is, in its formal structure, a behaviourist learning system. An agent interacts with an environment, receives rewards or punishments for its actions, and adjusts its policy to maximise cumulative reward [10]. The internal state of the agent is not a representation of knowledge in any structured sense; it is a set of parameters optimised to produce the right outputs given the right inputs. The correspondence extends to at least four fundamental characteristics, as shown in Table 1.

**Table 1** Behaviorism and Reinforcement Learning: Shared Structural Commitments

| **Behaviorism** | **Reinforcement Learning** |
|---|---|
| Learning is stimulus-response conditioning described through observable events. | Agents learn by receiving rewards or punishments for actions in an environment [10]. |
| Internal factors are excluded; only measurable outcomes matter. | RL maximises observable rewards rather than modelling internal cognitive processes [11]. |
| Humans and animals learn through the same environmental mechanisms. | RL is designed to mimic human and animal learning processes [10]. |
| The objective is habit formation through repeated conditioning. | Policy convergence through repeated interaction is analogous to habit formation [11]. |

What RL inherits from behaviourism is not merely a metaphor but a structural commitment: the exclusion of internal representational structure as a first-class object of the theory. The agent parameters encode statistical regularities in the reward landscape, not a structured understanding of why certain actions lead to certain outcomes. This is precisely the limitation that Chomsky identified in behaviourist accounts of language: the framework cannot distinguish between a system that has learned the rules and a system that has merely memorised the correct responses [5]. In the terms of Section 2: behaviourism bequeathed to RL an architecture that is structurally indifferent to the internal organisation of knowledge, and therefore structurally indifferent to systematicity.

## 3.2 Cognitivism and Deep Learning

Cognitivism emerged in the late 1950s as a direct response to the explanatory limits of behaviourism [12]. Where behaviourism treated the mind as a black box, cognitivism made the contents of that box the primary object of study. Learning, in the cognitivist view, is about how information is received, organised, stored, and retrieved by the mind [13]. The learner is active, not passive; new information is integrated with prior knowledge; and the organisation of mental representations is itself a determinant of learning outcomes [14].

The mapping to deep learning is immediate and pervasive. Neural networks are designed to transform, organise, and compress information through layered representations. Embedding techniques serve as models of knowledge representation [15]; Long Short-Term Memory networks model the retention and selective forgetting of sequential information [16]; attention mechanisms formalise the process of selective focus [17]; and meta-learning approaches operationalise the cognitivist notion of metacognition, learning how to learn [18].

**Table 2** Cognitivism and Deep Learning: Shared Structural Commitments

| **Cognitivism** | **Deep Learning** |
|---|---|
| Information is received, organised, stored, and retrieved by the mind. | Preprocessing, feature engineering, neural network layers, and embedding spaces serve analogous functions. |
| Mental representation: the mind represents and processes information internally. | Knowledge representation (embeddings), reasoning modules, and memory networks (e.g., LSTM) [16]. |
| Perception and attention are studied as core cognitive processes. | Attention mechanisms and perceptual models in computer vision [17]. |
| Metacognition: monitoring and controlling one's own cognitive processes [19]. | Meta-learning is implemented as learning-to-learn, inspired by cognitivist principles [18]. |

Yet deep learning inherits the central weakness of cognitivism: the opacity of internal representations. Cognitivism posited that mental representations are organised and retrievable, but struggled to specify how such representations could be observed or measured [20]. Deep learning faces the same challenge in amplified form. Recent work in mechanistic interpretability has made significant progress in identifying interpretable circuits within trained networks [65, 66], and probing classifiers can detect the presence of specific linguistic features in hidden representations [67]. These are important advances. However, they are diagnostic tools applied *after* training to representations that were not designed to be interpretable; they do not change the fact that the representations themselves are statistical compressions of the training distribution rather than organised knowledge structures that permit principled update, modular reuse, or transparent inspection. Similarly, neuro-symbolic approaches [56, 57] address compositionality by grafting symbolic reasoning modules onto neural substrates, but this typically produces a division of labour within an entangled system rather than a principled architectural separation. When a deep learning model fails, its failure still cannot be traced to a specific representational deficiency; the entire parameter space must be adjusted, often at the cost of what it previously knew, a phenomenon known as catastrophic forgetting [23]. The corrective techniques that have emerged to address these failures, such as interpretability methods, symbolic overlays, and retrieval augmentation, are themselves instances of the $P^*$ auxiliary hypotheses whose proliferation Section 2 identified as diagnostic of architectural limitation.

## 3.3 Constructivism and Emerging Integrative Approaches

Constructivism, the third major learning paradigm, holds that knowledge is actively constructed by the learner through engagement with the environment, building upon prior understanding [24]. It shares assumptions with both preceding paradigms, such as the active learner from cognitivism and the importance of environmental interaction from behaviourism, but adds a crucial principle: knowledge cannot be derived from nothing. New understanding must be built upon existing knowledge, and learning is always contextual [25].

In artificial intelligence, constructivist principles appear in curriculum learning, where models are trained on progressively more complex data [26]; in transfer learning, where knowledge from one domain is reused in another [27]; and in knowledge distillation, where the structured outputs of one model serve as a training signal for another [28]. More broadly, the entire movement toward compositional and modular approaches in AI reflects a constructivist intuition: that complex knowledge should be assembled from simpler, reusable components.

The limitation that constructivism inherits and passes to AI is the absence of a formal account of the construction process itself. Constructivism correctly insists that knowledge is built, not merely absorbed, but it does not specify the operators by which construction proceeds, the constraints under which it operates, or the structural properties that the resulting knowledge must satisfy. In AI, this translates to systems that can accumulate representations but cannot guarantee that those representations are compositionally structured, systematically recombinable, or resistant to interference when new knowledge is added.

## 3.4 Rote Learning: A Cross-Cultural Reappraisal

Beyond the three major paradigms, a fourth approach to learning deserves examination precisely because it has been systematically mischaracterised in the Western tradition and, consequently, underutilised in artificial intelligence. Rote learning, understood in the Western pedagogical tradition as the acquisition of information through mere repetition and memory without understanding [29], has been mapped to AI in its most minimal form: hard-coded rule systems and direct knowledge injection, where the programmer encodes knowledge, and the machine executes it without any representational engagement [30]. In this view, rote learning is deterministic, passive, and offers no path to generalisation.

However, the Eastern pedagogical tradition particularly Chinese and East Asian educational philosophy offers a fundamentally different interpretation. In Eastern culture, memorisation is not opposed to understanding; it is a structured phase within a broader learning process that leads toward understanding [29]. As Biggs [31] summarises the Chinese popular saying: repetition is the route to understanding. This is not a naive conflation of memory and comprehension. Rather, the Eastern view posits that systematic memorisation produces stable, retrievable representational fragments that serve as the necessary ground upon which understanding can later be constructed. Li [29] identifies five phases within this broader conception: repetition, memorisation, understanding, practice, and reviewing, a cycle in which memorisation is the first stage of an active learning process, not its terminus.

The distinction between the Western and Eastern conceptions maps onto a distinction directly relevant to AI architecture. The Western view treats memorisation as storage without structure, a flat database to be queried. The Eastern view treats memorisation as the construction of focused, context-specific representations under defined rules and constraints, representations that do not yet constitute understanding but that provide the organised prerequisites from which understanding can be built. The difference is between a bag of disconnected facts and a set of grouped puzzle pieces: the pieces do not yet form a picture, but their groupings accelerate the assembly process because they encode partial structural information.

This reappraisal suggests a reconceptualisation of rote learning for AI. Rather than equating it with deterministic programming, one can define it as a systematic approach by which a learner extracts and memorises information using rules and repetition in an active, constrained setting, what we may call a "game". The memorisation that results is characterised as the representation of facts or interactions without connections to other knowledge: isolated but structured fragments. This framing positions rote learning not as an alternative to more sophisticated learning paradigms but as a structured pre-learning phase that produces the representational prerequisites upon which constructivist or compositional learning can operate.

The implications for AI are twofold. First, the Western reduction of rote learning to passive storage has limited its role in AI to rule-based systems and lookup tables, missing the possibility that active, constrained memorisation could serve as a principled first stage in a multi-phase learning architecture. Second, the Eastern conception aligns naturally with the framework proposed in Section 4: the rote learning phase would correspond to the Memory component acquiring initial constraint signatures under the direction of a minimal Identity, producing the structured fragments that the Intellect subsequently decomposes and recomposes into genuine understanding. This connection between Eastern pedagogical philosophy and modular AI architecture represents, to the knowledge of the authors, a novel bridge that deserves further investigation.

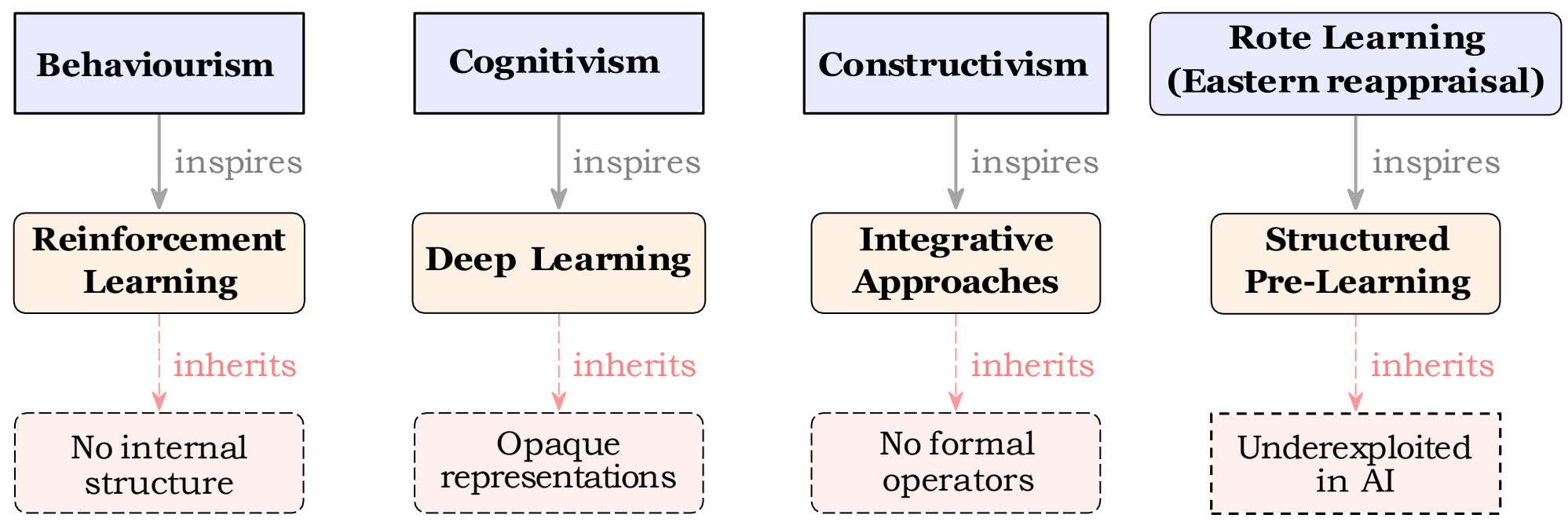

**Fig. 1** Genealogy from psychological learning theories to AI paradigms and their inherited structural limitations. Each psychological paradigm inspired a corresponding AI methodology, which in turn inherited the structural limitations of its psychological ancestor. The Eastern reappraisal of rote learning suggests a fourth, underexploited pathway.

## 3.5 The Diagnosis: Inherited Entanglement

The genealogy reveals a pattern that reinforcement learning, following behaviourism, cannot represent the internal structure of knowledge. Deep learning, following cognitivism, represents internal states but cannot organise them transparently. Constructivist-inspired approaches recognise that knowledge must be built compositionally but lack a formal account of how that construction produces representations that are necessarily systematic. And the Western interpretation of rote learning, by reducing memorisation to passive storage, has foreclosed the possibility that structured pre-learning could serve as a foundational phase in a more complete architecture.

They are manifestations of a single architectural choice: the entanglement of reasoning, purpose, and knowledge in a shared representational substrate. Reinforcement learning entangles the reward signal (Role, identity-alike) with the policy (reasoning) and the value estimates (memory or knowledge). Deep learning combines learned features (memory), the network architecture (reasoning), and the training objective. Each entanglement produces a system that works, often impressively, within the distribution it was trained on, but that cannot adapt outside it because the components cannot be updated independently. This is why the corrective techniques documented in Section 2 are necessary: the base architectures are structurally indifferent to systematicity because the precision at which stands the understanding of the psychological paradigms that shaped them never required it.

This diagnosis resonates with broader currents in the philosophy of cognitive science. The predictive processing framework of Clark [62, 63] argues that cognition is fundamentally a process of hierarchical prediction and error correction, in which the brain maintains generative models of its environment at multiple levels of abstraction. The predictive processing view supports the present argument in a specific way: it implies that the separation between what a system *knows* (its generative model), *how* it reasons (the prediction-error minimisation process), and *why* it attends to particular aspects of its environment (precision-weighting driven by goals and context) is not an optional design feature but a structural requirement for flexible, adaptive cognition. Current deep learning architectures collapse these distinctions. The taxonomy of creative cognition from Boden [64] offers a complementary perspective: she argues that transformational creativity, which involves altering the rules of a conceptual space rather than merely exploring it, requires the capacity to represent and manipulate the rules themselves independently of the content they generate.

After this diagnosis, the natural question is what architectural commitment would escape these inherited constraints?

# 4 The ReSynth Framework

The framework proposed here, ReSynth (Reasoning Synthesis), addresses the inherited limitations by separating the cognitive functions that current architectures entangle. The central claim is that reasoning, purpose, and knowledge are architecturally distinct functions that must be implemented as independent, modular components if the resulting system is to produce systematic behaviour by necessity rather than by accident. The architecture is illustrated in Fig. 2.

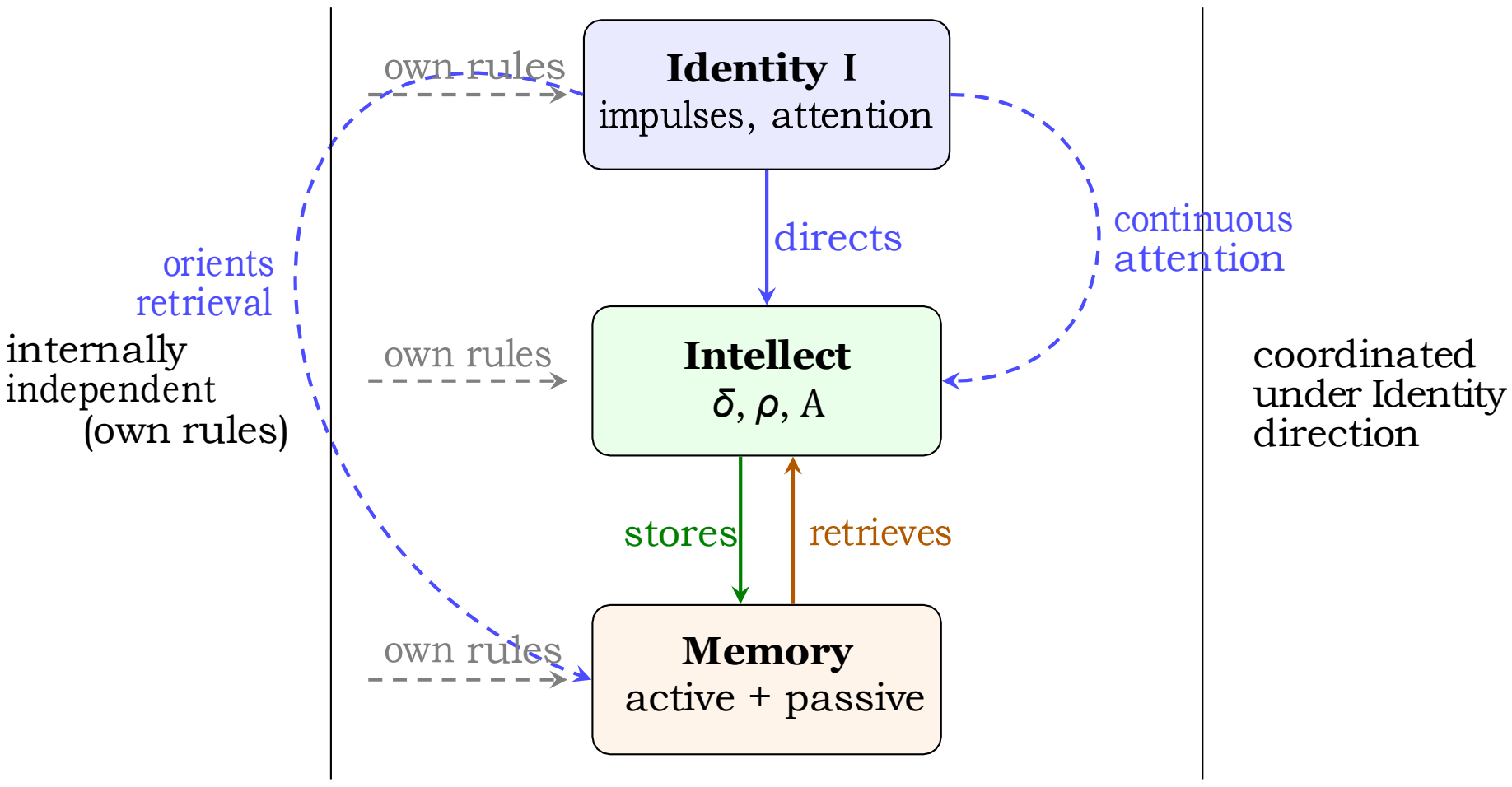

**Fig. 2** Independence and coordination in the trimodular architecture. Each component operates by its own internal rules (left), but the three interact as a coordinated system under the Identity's direction (right). The Identity provides both initial impulses and continuous attention throughout the reasoning cycle.

## 4.1 The Intellect

The Intellect is the reasoning engine of the framework. It is domain-agnostic: a universal component that applies two foundational operator types, decomposition and recomposition, to whatever phenomenon it encounters. Decomposition reveals the constraints that govern a structure, and recomposition assembles configurations that satisfy those constraints, including configurations that have never been directly observed. The Intellect carries no fixed identity, no predefined impulses, and no embedded purpose. It is the *how* of reasoning: the mechanism by which understanding is constructed.

This stands in direct contrast to the dominant paradigm, in which reasoning, representation, and purpose are blended in a single parameter space. When all three functions share the same weights, updating one necessarily affects the others. The Intellect independence means that the reasoning mechanism can be instantiated across different domains, given different purposes, and applied to different knowledge bases without modification. Recent work on minimal-data learning [43] supports the principle that reasoning does not require vast quantities of data but rather the capacity to apply a constrained set of operators to the phenomenon under observation.

Empirical support comes from recent research on modular architectures. A comprehensive review [44] demonstrated that modular neural networks consistently achieve superior compositional learning ability and sample efficiency compared to monolithic counterparts, precisely because modularity enables the learning and combination of atomic rules in novel ways, the compositional property that Fodor and Pylyshyn identified as essential for systematicity. Progressive neural networks [46], which are immune to forgetting through architectural separation, have been shown to outperform standard pretraining-and-finetuning baselines across reinforcement learning tasks. These findings provide indirect but substantial empirical support for a reasoning component that is architecturally separated from memory and purpose.

At a more formal level, the algebraic structure of the Intellect invites characterisation in the language of category theory [68, 69]. The decomposition and recomposition operators can be understood as morphisms within a category whose objects are constraint structures at varying levels of abstraction. The closure property (Definition 4 below) corresponds to the requirement that the category be closed under composition. A full categorical formalisation is beyond the scope of this paper, but the existence of a natural categorical framework provides confidence that the semi-formal characterisation given below can, in principle, be made fully rigorous.

### 4.1.1 Operators and Algebraic Structure

To make the architectural claims precise, this section provides a semi-formal characterisation of the operators of the Intellect. A full formalisation, including the algebraic specification and its application to syntactic parsing as a test domain, is the subject of ongoing work and will be reported in a dedicated paper.

**Definition 1** (Phenomenon and Scope) A **phenomenon** $P$ is any observable structure presented to the Intellect for analysis. A **scope condition** $E$, supplied by the Identity, specifies which aspects of $P$ the Intellect is to attend to and at what level of abstraction. The scope condition is external to the Intellect's own theory: it determines *what* the Intellect operates on, but does not alter *how* it operates.

**Definition 2** (Decomposition) The **decomposition operator** $\delta$ takes a phenomenon $P$ and a scope condition $E$ and produces a constraint set:

$$\delta : P \times E \longrightarrow \mathrm{C} = \{c_1, c_2, \ldots, c_n\},$$

where each $c_i$ is a constraint governing the structure of $P$ as observed under scope $E$. Constraints are relational: each $c_i$ specifies a necessary condition that any configuration of the phenomenon's components must satisfy. The decomposition does not produce the components themselves; it produces the *rules* that govern their arrangement.

**Definition 3** (Recomposition) The **recomposition operator** $\rho$ takes a subset of constraints and an algebra A and produces the set of all configurations that satisfy those constraints under A:

$$\rho : 2^{\mathrm{C}} \times \mathrm{A} \longrightarrow \mathrm{K},$$

where K is the set of valid configurations. Crucially, K may contain configurations not present in the original phenomenon $P$: the recomposition operator generates all structures permitted by the discovered constraints, including those never directly observed.

**Definition 4** (Closure) The operator pair $(\delta, \rho)$ is **closed under** A if and only if:

1. Every configuration $k \in \mathrm{K}$ produced by $\rho$ is itself a valid input to $\delta$: for any scope condition $E'$, $\delta(k, E')$ is defined and produces a constraint set.
2. Every constraint set $\mathrm{C}'$ produced by $\delta$ is a valid input to $\rho$: $\rho(\mathrm{C}', \mathrm{A})$ is defined and produces a configuration set.

Closure ensures that the reasoning process is indefinitely iterable: the outputs of decomposition can always be recomposed, and the outputs of recomposition can always be further decomposed.

**Proposition 1** (Systematicity as Structural Consequence) *If the operators of the Intellect* $(\delta, \rho)$ *are closed under algebra* A, *and if decomposition* $\delta(P, E)$ *produces a constraint set* C, *then* every *configuration in* $\rho(\mathrm{C}', \mathrm{A})$ *for any* $\mathrm{C}' \subseteq \mathrm{C}$ *is producible by the Intellect, including configurations not present in* $P$. *The capacity to analyse a phenomenon entails the capacity to generate all of its structurally permissible variants. Systematicity is a consequence of closure, not an auxiliary hypothesis.*

**Relation to Aizawa's analysis.** The critical distinction between this architecture and the connectionist (or classicist) frameworks that Aizawa analysed is the following. In connectionism, the base theory (nodes with weighted connections) does not entail systematicity; producing systematic outputs requires an auxiliary property $P^*$ that the base theory does not necessitate. In ReSynth, the base theory of the Intellect *is* the closed algebra $(\delta, \rho, \mathrm{A})$. The scope condition $E$, provided by the Identity, is *external* to this theory: it determines what the Intellect operates on, not how it operates. This is analogous to the distinction Aizawa draws between the Copernican and Ptolemaic cases. The closed algebra produces systematic outputs for *whatever* scope it is given, just as the Copernican model produces bounded elongation for *whatever* inner planet it is applied to (see Fig. 3).

## 4.2 The Identity

The Identity is the component that provides purpose and direction. It defines *why* the Intellect reasons and *what* it attends to. The Identity carries impulses, the drives that initiate and orient the reasoning process and provides continuous attention, the mechanism by which those impulses orient the Intellect toward specific constraints and specific levels of abstraction. The Identity is not a role, and does not generate reasoning; it directs its trajectory.

This directive function addresses a foundational problem in cognitive architecture that has remained open since at least the systematicity debate of Fodor and Pylyshyn. Aizawa [39] demonstrated that neither Connectionism nor Classicism can explain the systematicity of thought: connectionist networks are architecturally indifferent between producing systematic and non-systematic representations, while classical compositional structure alone does not entail systematicity either, a Turing machine with a compositional language of thought can be programmed to produce *any* subset of well-formed formulae, systematic or not. Both frameworks require additional hypotheses that, within their own architectures, remain arbitrary in precisely the way that Ptolemaic colinearity hypotheses were arbitrary relative to geocentrism. What is missing, on the analysis of Aizawa, is a theoretical commitment from which systematicity follows *necessarily* rather than as one option among many. The Identity, as formalised here, supplies exactly this: a fixed directive constraint that determines which representational trajectories the Intellect pursues, rendering the resulting systematicity non-arbitrary with respect to the architecture.

Recent work within the connectionist paradigm provides striking independent support for this principle. The Semantic Tube Prediction (STP) framework [52], a JEPA-style regulariser for language models, demonstrates that confining hidden-state trajectories to a tubular neighbourhood of a semantic geodesic achieves 16× greater data efficiency than unconstrained scaling, directly violating Chinchilla-style scaling laws. The geometric prior that drives the efficiency of the STP functions as precisely the kind of non-arbitrary architectural commitment whose absence Aizawa [39] identified in both classical and connectionist accounts of systematicity. Aizawa showed that the bare hypothesis of nodes and weighted connections is "absolutely indifferent" between producing systematic and non-systematic representations, and that even augmenting the architecture with an additional property $P*$ merely replicates the Ptolemaic astronomer's colinearity hypothesis, a post-hoc stipulation that saves the phenomena without explaining them. What the STP framework suggests is that the relevant explanatory work is performed not by representational capacity (which, as Aizawa demonstrated, is architecturally indifferent to systematicity) but by a *directive constraint* that channels that capacity along non-arbitrary trajectories: direction, not scale.

This convergence is not accidental. The line of research of LeCun for autonomous machine intelligence [21] grounds JEPA in an embodied cognition commitment: learning world models through latent prediction shaped by sensorimotor engagement rather than disembodied symbol manipulation [22]. STP demonstrates that this programme succeeds precisely when a directive geometric constraint channels the latent dynamics. From the perspective of the present research, embodied cognition offers one of the viable approaches to bridging the disciplinary chasm between identity theory in AI and intelligence research. Identity, on this account, functions as the persistent attentional bias that determines which latent trajectories are salient, supplying the orientational commitment that a given JEPA instantiation implements but does not, *qua* architecture, explain. Just as Aizawa argued that compositionality does not entail systematicity, and that even recursivity fails to guarantee the specific systematic relations actually observed in cognition, the JEPA architecture does not entail any particular directive orientation. Identity is the candidate for the non-arbitrary constraint that the analysis of Aizawa showed both paradigms lacked.

The Identity in ReSynth tries to generalise this insight architecturally. Where STP imposes a fixed geometric prior within a monolithic architecture, the Identity provides a modular, reconfigurable source of direction that can be swapped across domains without altering the reasoning engine itself. The same Intellect, given a syntactician Identity, follows trajectories oriented toward syntactic constraints; given a semanticist Identity, it follows trajectories oriented toward referential and entailment constraints. The operators do not change. The direction does.

Thus, the Identity is not a post-hoc correction. It does not filter outputs after they are generated, as alignment techniques do in current models [42]. It operates upstream: it determines which constraints the Intellect attends to, which operators it prioritises, and what knowledge structure results. This is alignment by design, not alignment by correction. Current alignment methods through RLHF have been shown to reduce surface-level toxicity and improve instruction-following, but they do not alter the underlying representational structure, leaving models vulnerable to systematic reasoning failures under distribution shift [37].

The significance of this component can be illustrated through the history of language itself. Language is, in its essence, a reasoning interface, a system of decomposition and recomposition applied to shared experience. But the specific characteristics of human language were shaped by the purposes that drove its creation: not merely survival, but the social needs to negotiate, persuade, express solidarity, and deceive [49, 50]. The Identity is the generative force that determines which knowledge the Intellect constructs and how that knowledge is structured. A full treatment of the Identity component will be presented in a dedicated paper.

## 4.3 Memory

Memory in ReSynth is not a store of raw data or a compressed parameter space. It is a structured map of constraint-to-operator signatures: records of which constraints were detected, which operators were applied, and the resulting understanding. Two types of memory are distinguished. Active memory is small, focused, and holds the current state of a reasoning process. Passive memory is expansive and serves to accelerate reasoning by surfacing relevant prior signatures when the Intellect encounters a new phenomenon.

This architecture contrasts sharply with the memory function of current deep learning models, which store knowledge as compressed statistical correlations. The consequence is that current models can retrieve patterns resembling what they have seen but might struggle to construct new decompositions of what they have not [47, 48]. Empirical evidence shows that LLM factual accuracy degrades sharply for entities that appear infrequently in training data, with the factual precision of GPT-4 dropping below 40% for low-frequency entities compared to over 90% for popular ones [37, 45]. The reappraised conception of rote learning integrates naturally with this architecture. The Eastern five-phase cycle: repetition, memorisation, understanding, practice, reviewing, maps onto the interaction between Memory and the Intellect. The first two phases correspond to the acquisition of initial constraint signatures in passive memory. The third phase (understanding) corresponds to the Intellect's decomposition and recomposition of those fragments into genuine constraint-to-operator mappings. The final phases correspond to consolidation and refinement through repeated application of practice and review. The Eastern conception of rote learning anticipates the architectural separation between structured memorisation and active reasoning that the current framework formalises.

## 4.4 The Separation Thesis

The core architectural claim of ReSynth is that the separation of Intellect, Identity, and Memory is not merely a design preference but a structural requirement for systematic behaviour. When these functions are merged in shared parameter spaces, the system cannot update one without affecting the others. New knowledge interferes with existing knowledge (catastrophic forgetting or hallucination), the reasoning process cannot be inspected or transferred independently of the knowledge it was trained on, and the purpose of the system must be imposed externally through corrective mechanisms rather than specified architecturally.

By separating these components, ReSynth ensures that the operators of the Intellect remain universal, that the Identity can be reconfigured without affecting stored knowledge, and that Memory can be extended additively without degrading the reasoning mechanism. Adaptation is a consequence of the fact that each component can be updated independently while the others remain stable.

However, independence must be understood precisely. Each component is internally autonomous: the Intellect has its own operators and algebraic rules, Memory has its own storage and retrieval dynamics, and the Identity has its own impulses and attention logic. None depends on the internal machinery of the others to function. But this internal independence does not mean isolation. The three components operate as a coordinated system under the direction of the Identity. Independence is a property of how each component is built; coordination is a property of how they interact under the guidance of the Identity.

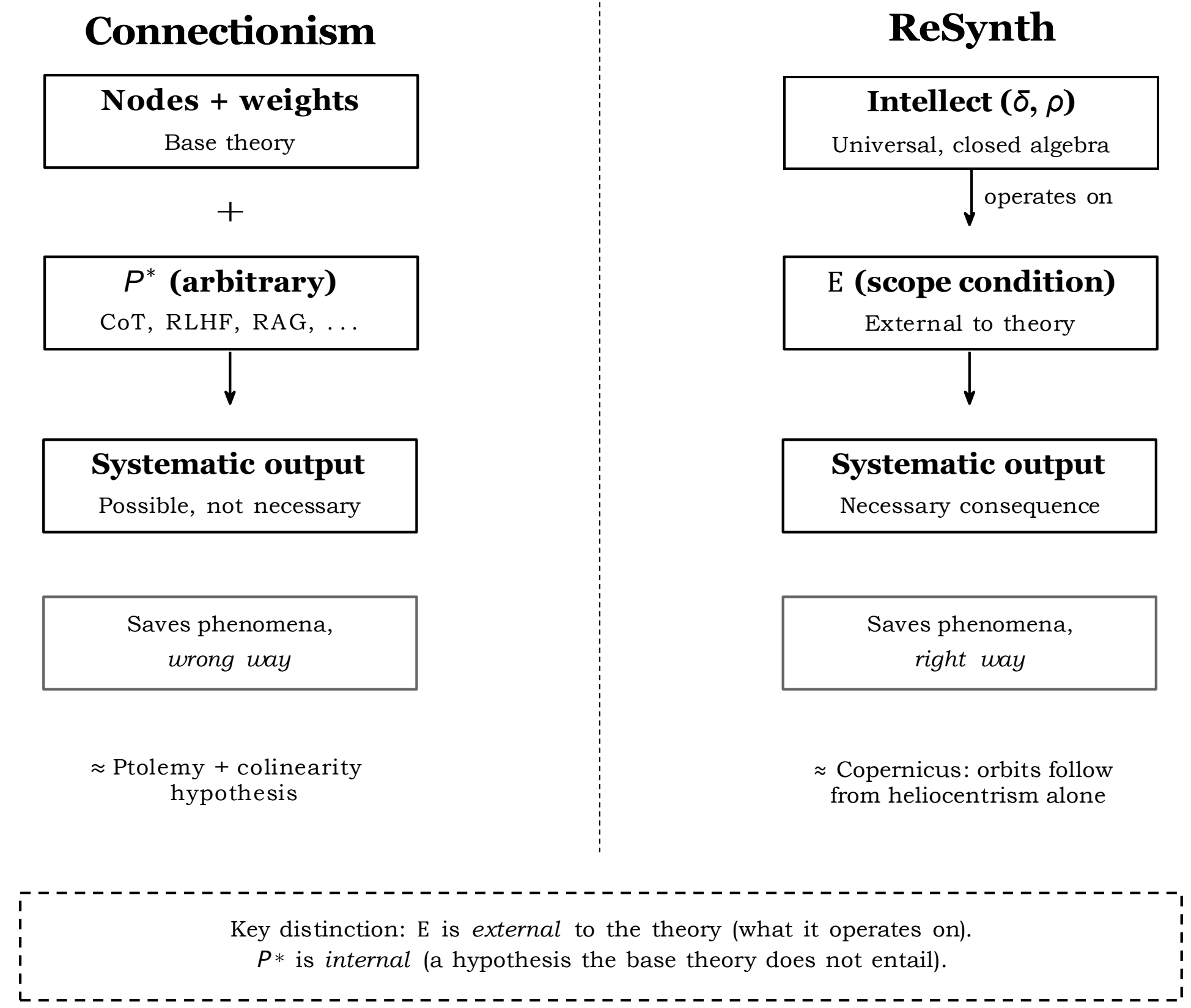

**Fig. 3** Addressing the Ptolemaic Objection. **Left:** Connectionism requires arbitrary auxiliary hypotheses ($P*$) conjoined with the base theory to produce systematic behavior. **Right:** In ReSynth, systematic output is a necessary consequence of the closed algebra of the Intellect operating on an external scope condition $E$.

## 4.5 ReSynth in Relation to Existing Architectures

The claim that the separation of reasoning, purpose, and memory is architecturally necessary invites comparison with existing systems that have pursued related forms of modularity. Five frameworks are particularly relevant.

### *ACT-R.*

Anderson's ACT-R [53] separates declarative memory from procedural memory and mediates their interaction through buffers including a goal buffer. In several respects, ACT-R anticipates the trimodular structure of ReSynth. However, ACT-R's production rules are *domain-specific*, and its goal buffer is part of the same production system that performs reasoning, not an independent module with its own internal rules. The difference is between modularity of *representation* (ACT-R separates what is stored) and modularity of *function* (ReSynth separates how reasoning, direction, and storage operate).

### *Soar.*

Laird's Soar [54] uses a unified processing cycle with long-term memory divided into procedural, semantic, and episodic stores. Soar's central innovation is *impasse-driven learning*. However, Soar entangles purpose with reasoning: the goal stack is *generated by the reasoning process itself* through impasse creation, not provided by an independent directive component. A Soar agent that has learned to process one configuration has no architectural guarantee of being able to process its systematic variants.

### *DreamCoder.*

Ellis et al.'s DreamCoder [55] is the closest in spirit to ReSynth. Its library corresponds to Memory, its search procedure to the Intellect, and its task distribution to the Identity. However, three differences distinguish it from ReSynth: the library and search procedure are *co-adapted* during the sleep phase; there is no explicit Identity component; and its operators are those of a typed functional programming language rather than pre-formal constraint operators.

### *Neuro-symbolic approaches.*

Neuro-symbolic systems [56, 57, 75] such as Neural Theorem Provers [72], DeepProbLog [73], and Scallop [74] address compositionality by combining neural networks with symbolic reasoning. However, from the perspective of the separation thesis, their components remain representationally entangled: updating symbolic rules may require retraining the neural grounding, and vice versa. In Aizawa's terms, the symbolic component functions as a sophisticated $P^*$.

### *Global Workspace Theory.*

Baars' GWT [58, 59] and its computational descendants [60, 61] propose modular processors coordinated by a global workspace. The key difference from ReSynth is that GWT does not distinguish between purpose and reasoning; its workspace is a coordination mechanism, not a persistent directive source.

### *Summary.*

Table 3 summarises the comparison. The common thread is that existing architectures achieve modularity along one or two dimensions but not all three. None achieves the *tripartite* separation of reasoning from purpose from memory that ReSynth proposes as the structural requirement for systematic behaviour.

What ReSynth adds is the claim that these partial separations are insufficient: only when reasoning, purpose, and memory are *constitutively* independent, each operating by its own internal rules, coordinated but not entangled, does systematicity become a structural consequence rather than an accidental product of auxiliary hypotheses.

**Table 3** Comparison of cognitive and AI architectures along three dimensions of separation.

| **Architecture** | **Reasoning** | **Purpose / Direction** | **Memory / Knowledge** |
|---|---|---|---|
| ACT-R [53] | Domain-specific productions | Goal buffer (coupled to productions) | Declarative / procedural split |
| Soar [54] | Task-specific operators | Goal stack (generated by reasoning) | Procedural / semantic / episodic |
| DreamCoder [55] | Neural-guided search | Implicit (task distribution) | Library (co-adapted) |
| Neuro-symb. [56] | Symbolic + neural (entangled) | Training objective (shared) | Distributed parameters |
| GWT [58] | Specialised modules | Workspace (coordination only) | Distributed across modules |
| **ReSynth** | **Universal** $(\delta, \rho, \mathrm{A})$ | **Independent Identity** | **Structured signatures** |

# 5 Conclusion

This paper has argued, in three stages, that the dominant failure mode of current AI, the inability to systematically recombine known components in novel configurations, has architectural origins that trace back to the psychological theories from which AI paradigms were derived.

The first stage established the problem through the systematicity debate: drawing on the demonstration of Aizawa that neither connectionism nor classicism can make systematicity an architectural necessity, it showed that the corrective techniques proliferating in modern AI function as auxiliary hypotheses ($P^*$) that address symptoms without resolving the underlying architectural indifference.

The second stage explained the origin: a genealogy from behaviourism, cognitivism, and constructivism to reinforcement learning, deep learning, and integrative approaches, in which each AI paradigm inherited a specific structural limitation from its psychological ancestor. A cross-cultural reappraisal of rote learning revealed a further underexploited pathway: the Eastern conception of memorisation as a structured, multiphase precursor to understanding offers a more adequate model for how initial representations should be acquired.

The third stage proposed an escape: the ReSynth framework, which separates reasoning (the Intellect), purpose (the Identity), and memory into constitutively independent components. Through semi-formal characterisation of the operators of the Intellect and their closure property, the paper argued that systematic behaviour becomes a structural consequence of this architecture rather than a correction applied after the fact.

Yet the practical success of systems that avoid rather than achieve systematicity raises a question this paper has deliberately left open: whether systematicity is, in fact, a necessary condition for the kinds of intelligence that matter most. The overwhelming majority of deployed AI systems solve their target problems without any architectural guarantee of systematic recombination, and their commercial and scientific utility is not in dispute. If the goal is narrow prediction, next-token, next-frame, next-reward, then the auxiliary-hypothesis strategy may not be a failure mode, but a perfectly adequate engineering trade-off. The argument of this paper bites only if one accepts that the problems of society will increasingly need AI to address open-ended adaptation, cross-domain transfer, robust reasoning under distributional shift which are problems for which systematicity is constitutive rather than incidental. That assumption has been operative throughout, but it is an empirical bet, not a settled fact.

If, however, that bet is correct, if the next generation of problems does demand the kind of unbounded, non-arbitrary recombination that Aizawa showed that neither classical nor connectionist architectures can guarantee, then the Identity component of the ReSynth framework ceases to be a theoretical nicety and becomes an architectural necessity.

Systematicity, as this paper has argued, does not emerge from representational capacity alone; it requires a directive constraint that channels that capacity along non-arbitrary trajectories. The Intellect supplies the compositional machinery, and memory supplies the substrate, but neither specifies *which* recombinations matter, *which* abstractions to preserve across contexts, or *which* distributional shifts to treat as salient rather than noise. That orientational work deciding what counts as the same problem in a new guise, is precisely the function this framework assigns to Identity: the persistent evaluative bias that transforms an architecture indifferent to systematicity into one that exhibits it as a structural consequence. A rigorous programme of research on Identity theory in AI, understood not as a social category but as the computational locus of directive commitment, therefore becomes the critical path. Determining whether this commitment can be formalised with the same precision as the closure of Intellect operators and whether the resulting architecture delivers systematic behaviour where current systems fail, is the task that the present theoretical groundwork is designed to make tractable, and the subject of ongoing empirical work to be reported separately.

## Statements and Declarations

**Funding.** The work was done with partial support from the Mexican Government through the grant A1-S47854 of CONACYT, Mexico, grants 20250738, 20260367, of the Secretaría de Investigación y Posgrado of the Instituto Politécnico Nacional, Mexico. The authors acknowledge the support of Microsoft through the Microsoft Latin America PhD Award.

**Competing Interests.** The authors declare no competing interests.

**Author Contributions.** Alex Anvi Eponon: Conceptualisation, Investigation, Methodology, Writing – original draft, Writing – review & editing. Ildar Batyrshin: Supervision, Writing – review & editing. Christian E. Maldonado-Sifuentes: Writing – review & editing. Grigori Sidorov: Supervision, Writing – review & editing.

**Data Availability.** No datasets were generated or analysed during the current study.

**Acknowledgments.** The authors thank CONACYT for the computing resources brought to them through the Plataforma de Aprendizaje Profundo para Tecnolog´ıas del Lenguaje of the Laboratorio de Supercõmputo of the INAOE, Mexico.